\documentclass{article}

\usepackage{microtype}
\usepackage{graphicx}
\usepackage{subfigure}
\usepackage{booktabs} 
\newcommand\given[1][]{\:#1\vert\:}
\usepackage{amssymb}
\usepackage{amsmath}
\usepackage[]{algorithm2e}
\usepackage{subfiles}
\usepackage{setspace}
\usepackage{hyperref}

\usepackage[accepted]{icml2020}

\icmltitlerunning{Student-Teacher Curriculum Learning via Reinforcement Learning}

\begin{document}

\twocolumn[
\icmltitle{Student-Teacher Curriculum Learning via Reinforcement Learning: \\
            Predicting Hospital Inpatient Admission Location}



\icmlsetsymbol{equal}{*}

\begin{icmlauthorlist}
\icmlauthor{Rasheed el-Bouri}{UoO}
\icmlauthor{David Eyre}{UoO,OUH}
\icmlauthor{Peter Watkinson}{OUH}
\icmlauthor{Tingting Zhu}{UoO}
\icmlauthor{David A. Clifton}{UoO}

\end{icmlauthorlist}
\icmlaffiliation{UoO}{University of Oxford}
\icmlaffiliation{OUH}{Oxford University Hospitals Trust}


\icmlcorrespondingauthor{Rasheed el-Bouri}{rasheed.el-bouri@eng.ox.ac.uk}

\icmlkeywords{Machine Learning, ICML}

\vskip 0.3in
]



\printAffiliationsAndNotice{}  

\begin{abstract}
    Accurate and reliable prediction of hospital admission location is important due to resource-constraints and space availability in a clinical setting, particularly when dealing with patients who come from the emergency department. In this work we propose a student-teacher network via reinforcement learning to deal with this specific problem. A representation of the weights of the student network is treated as the state and is fed as an input to the teacher network. The teacher network's action is to select the most appropriate batch of data to train the student network on from a training set sorted according to entropy. By validating on three datasets, not only do we show that our approach outperforms state-of-the-art methods on tabular data and performs competitively on image recognition, but also that novel curricula are learned by the teacher network. We demonstrate experimentally that the teacher network can actively learn about the student network and guide it to achieve better performance than if trained alone.
\end{abstract}
\section{Introduction}
\label{Introduction}
A major problem that many hospital wards face is the timely preparation of beds for patients from the emergency department (ED) who are due to be admitted as inpatients \cite{staib2017uniting}. The average time between initial examination (triage) in the ED and a request for a bed in the hospital is 4 hours \cite{francis2013report} and the only information available to staff about the patient is that collected at triage. This information can be helpful to deduce the patient's condition, which in turn can be used to predict which of the relevant inpatient wards needs to prepare space. Often, the ward is requested after the 4 hours have passed in a reactive manner. Predictively allocating patients to wards however, should allow for the timely care and admission of the patient as well as allocation of resource to the relevant departments. Furthermore this may contribute to reducing crowding of the ED as well as patient waiting times. With this in mind, we present a method of predicting where in a hospital emergency patients will be admitted after being triaged using a combination of reinforcement and curriculum learning. 

Curriculum learning was introduced to train a neural network in a similar manner to that in which humans are educated. The approach stems from the observation that children in schools learn by beginning with simple ideas and progressing on to more complex topics; it is believed that neural networks may also benefit from this structured approach to learning. By initially presenting the network with data that are `easier' to fit, the optimisation surface (of network prediction error vs. network parameters) is more likely to be convex \cite{bengio2009curriculum}. 

\par In this paper we propose using neural networks for classification by exploiting this notion of a curriculum and extending the teacher-student curriculum learning method of \cite{matiisen2017teacher}. We use two networks, where one (the teacher) is trained with information about the state of the other (the student) in order to guide training and maximise performance. A teacher's observations of many previous students provides them with an idea of how a student can best learn \cite{papay2015productivity} using a pre-defined curriculum. This work aims to mimic this action of learning from an agent (that is itself learning), with the aim of creating a policy (the function that maps a state to an action) to guide a future agent to achieve successful performance.
\par The contributions of this work include the development of a student-teacher learner, where the teacher is trained using reinforcement learning which, to the best of the authors' knowledge, is the first work to propose such a student-teacher configuration. It is also believed to be the first work to use a representation of the weights extracted from the student as inputs to the teacher. This method has demonstrably better performance for our healthcare application than all others to which it was compared. We believe that the resulting classifier will prove a useful tool for hospitals in order to predict the resources required ahead of time and improve the movement of patients out of the ED and into the wards. \par Section \ref{section:relatedwork} discusses related work and Section \ref{section:methodology} outlines the setup of the student-teacher network. Results are in Section \ref{section:results} and we discuss the curricula generated by the teachers in Section \ref{section:discussion} before outlining the limitations of this approach in Section \ref{section:future_work}.

\section{Related Work}
\label{section:relatedwork}

In this work, we create a teacher agent that learns to train a feedforward student network by selecting the data to present to the student. We organise our data into batches according to a curriculum. We define the `easiness' of a batch of data according to the entropy within the batch; i.e, a batch $B_i$ is said to be easier than batch $B_j$ if $H \lbrack B_i \rbrack < H \lbrack B_j \rbrack$, where $H$ is the entropy of the batch. It has been shown in \cite{ wu2016training} and \cite{pentina2015curriculum} amongst others that training neural networks in a structured fashion by presenting batches from easiest to hardest leads to improvement in performance of the network, as well as increasing the speed of training. However, we also see in the works of \cite{hacohen2019power} and \cite{bengio2009curriculum} that training in an `anti-curriculum' (hardest to easiest data) can also lead to improved performance. A curriculum may also be defined for specific problem sets, with results from \cite{zaremba2014learning} and \cite{bengio2015scheduled} showing that a curriculum tailored for the task at hand provided better results than a general curriculum. This is indicative that training according to some type of structure is important for learning, and that structure may be task specific.
\par Student-teacher curriculum learning has been explored in \cite{matiisen2017teacher} where a similar initial configuration to that proposed here is used. Progress of the student is measured through its improvement in task performance, which in turn affects the probability of selecting a curriculum batch on which to train. The authors proposed four different algorithms to monitor the progress of the student. These are (i) the use of a non-stationary bandit to select the next training batch, (ii)-(iii) using linear regression and a windowed linear regression on the task accuracy to predict the batch most likely to provide the greatest improvement in performance, and (iv) using Thompson sampling to select the next batch. While these approaches have an effective performance on those specific problems presented in the existing studies, the selection of data is not based on information pertaining to the state of the student (but instead is based on prior information of the student in its previous state, and from previous performance of the algorithms). In this work, we propose using knowledge of the current state of the student to develop a policy for the teacher to train future students. Another work which uses a similar approach is that of \cite{graves2017automated} where a curriculum is also generated and a non-stationary bandit is used with the EXP3.S algorithm \cite{auer2002nonstochastic}. However, again ignoring the state of the student, this latter algorithm is conditioned on the previous performance of the student.
\par Reinforcement learning has proven to be highly effective in many applications, canonically within the domain of games \cite{oh2015action,kelly2017multi}. The training of a neural network can also be seen as a game where the task is to choose the right combination of training data to maximise a score, which in this case is the prediction accuracy. This makes reinforcement learning an attractive choice for training the teacher network. Our approach has parallels with the field of hierarchical learning, in which a higher-level agent guides the training of a lower-level agent \cite{kulkarni2016hierarchical}. However to the best of the authors' knowledge, none of the methods used in hierarchical learning operate directly on the weight space of the lower-level agent.

\section{Methodology}
\label{section:methodology}
\subsection{Data Pre-Processing}
\label{section:pre-processing}
\par The first task for our model is to organise data into batches according to their complexity. As previously mentioned this can be achieved according to the entropy of the data (among other methods not described here for brevity). For tabular data (categorical and numerical data such as the information recorded at triage), we choose to represent the entropy through the use of the Mahalanobis distance. The Mahalanobis distance is defined as:
\begin{equation}
d\left({\mathbf{x}_n}\right) = \left(\left(\mathbf{x_n} - \boldsymbol{\mu} \right)^T \mathbf{S^{-1 }}\left({\mathbf{x_n} - \boldsymbol{\mu}}\right)\right)^\frac{1}{2}
\label{eq:Mahalanobis_distance}
\end{equation}
where $\mathbf{x_n}$ are the input features of datapoint $n$, $\boldsymbol{\mu}$ is the vector of the mean value of each feature, and $\mathbf{S}$ is the covariance matrix. As certain inputs to a neural network may be of different data types we encode the input data prior to organising into a curriculum using a denoising autoencoder for the Mahalanobis distance to be defined in a scalar space.

\par Let $\mathcal{D}$ be a training dataset consisting of inputs and  labels $x_n$ and $y_n$ respectively. We first train a denoising autoencoder such that $\phi \left(x_n\right) \approx x_n$. To construct our curriculum from this latent representation, we calculate the Mahalanobis distance, $d\left(\phi\left(x_n\right)\right)$ for all $n$. We then sort the data from the lowest Mahalanobis distance to the highest and create $N$ batches by dividing the whole set into $N$ separate batches. Two methods of creating the batches were investigated:
\begin{enumerate}
    \item {All batches are separate and contain separate training data; i.e, if $B_0$ is the batch with the lowest $d\left(\phi\left(x_n\right)\right)$ and $B_N$ that with the highest, then $B_0 \cap B_1 \cap \dots B_N = \varnothing$}
    \item {Batches are cumulative supersets; i.e, $B_0 \subset B_1 \subset \dots B_N$.}
\end{enumerate}

\begin{figure}[h]
\vskip 0.2in
\begin{center}
\centerline{\includegraphics[width=\columnwidth]{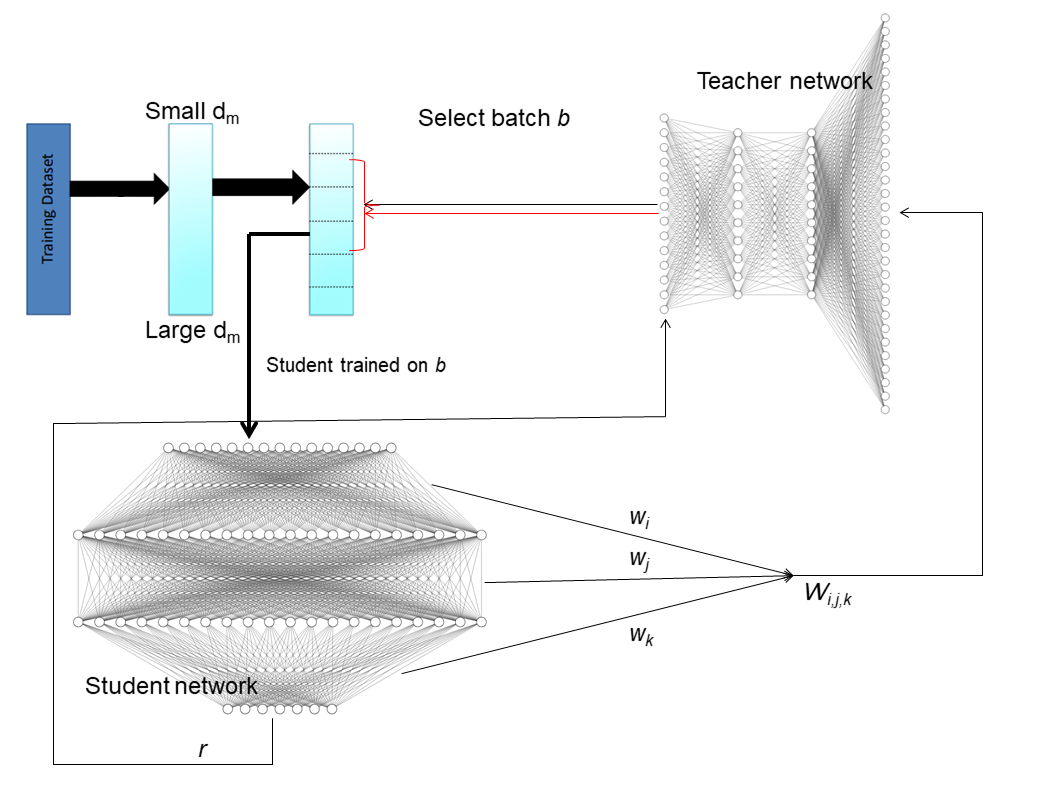}}\caption{The architecture of the model we are proposing. The data is separated according to $d_m$, the Mahalanobis distance. The reward from the student, $r$ is fed back to the teacher, as well as a representation of the weights from each layer of the student network. The first output of the teacher (black) is where along the sorted data to sample from, the second (red) is how much data around this to include in the batch.}
\label{fig:teacher_student_visual}
\end{center}
\vskip -0.2in
\end{figure}

\subsection{Teacher Network}
\label{section:teacher}
Figure \ref{fig:teacher_student_visual} shows the overall architecture for the student-teacher network being proposed in this work. The teacher network, as most reinforcement learning agents, is characterised by the state, action, reward, next state tuple (\textit{s,a,r,s'}) \cite{sutton1998introduction}. Using these we are able to discern the value of the student being in a given state and taking a certain action (being trained on certain data) for a given task. 
\par We seek to discover a curriculum that will improve the performance of the student over alternative training methods. As previously mentioned, we aim to condition on the current configuration of the student and so we define the \textit{state} therefore to be the weights of the student network, i.e, a representation of the weights between all of the nodes for all layers as described in Section \ref{section:student}.
\par The \textit{actions} of the teacher are selecting the ideal batch of training data that are organised by a curriculum, to train the student network. This is done through the policy by selecting the appropriate data to train with based on the output as described in Section \ref{section:teacher_action}. Note that in this problem, the number of actions can change depending on how many batches we define our curriculum to have, or the method with which we select data to create a batch. 
\par The \textit{reward} is given by the improvement in accuracy of the student compared to the previous accuracy achieved before the training data batch was trained, i.e., $r_t = \delta_{t} - \delta_{t-1}$ at step $t$ where $\delta$ is the accuracy of the student on the training set. However, this would lead to overfitting if only trying to maximise the training accuracy. As a result we measure the students accuracy on the training set and the validation set jointly by multiplying them as $\left(\delta_{t} - \delta_{t-1}\right)A_v$ where $A_v$ is the validation accuracy to formulate the new reward. This way if overfitting occurs, the validation accuracy reduces thereby reducing the overall reward. 
\par Our \textit{next state}, $\boldsymbol{s'}$ is the updated weights of the student network after training on the batch chosen by the teacher.
\par To train the teacher we must make use of the Bellman equation. This describes the value of an agent (any model or actor which has a state and the capability to take actions) being in a certain state for a given task. It can also describe the value of an agent being in a certain state and taking a certain action for that same task (Q-value) \cite{wiering2012reinforcement}. The equation is described by: \begin{equation}
    Q^{*}(s,a) = \mathbb{E}_{s'}\lbrack r + \gamma\max_{a'}Q^{*}(s',a')\given s,a\rbrack
    \label{Bellman_optimality}
\end{equation} where $Q^{*}$ is the optimal action-value function, $r$ is the reward of the action that was taken, $\gamma$ is the discount factor, $s,a$ and $s',a'$ are the current state and action and the next state and action respectively. The discount factor is a way of weighting whether future or immediate rewards are important, with high values of $\gamma$ favouring a long-term reward and low values favouring immediate reward. The discount factor is generally treated as a hyper-parameter. 
\par The intuition behind Equation \ref{Bellman_optimality} is that, if we have a function $Q^{*}$ which tells us the value of being in a certain state and taking a certain action, then naturally, optimal behaviour is given by choosing the next action which maximises this.

For large state spaces such as the weight space of the student, a function for the Q-values must be approximated \cite{gaskett1999q}. This can be done using function approximation methods such as deep neural networks.
Now that $Q(s,a)$ is dependent upon network parameters, $\theta$, it becomes $Q(s,a;\theta)$ (note that $Q$ here is the action-value function for the current policy and not $Q^*$, the optimal policy) and the loss function to train our teacher network is given by:
\begin{equation}
    \label{eq:Bellman_loss}
    \mathcal{L}(\theta_i) = \left(r + \gamma\max_{a'}Q(s',a';\theta^{-}_i) - Q(s,a;\theta_i) \right)^2
\end{equation}
where $\theta^{-}$ are the parameters of the target network (the version of the teacher that is held constant for $K$ steps to stabilise training as described in \cite{mnih2015human}). 
\par Experience replay is used to decorrelate the data being trained on with the current policy. This is required to stabilise the learning of the Q-function. A buffer, $R$ is kept of the previous $(s,a,r,s')$ tuples that the teacher has encountered. A random sample of $R$ of size $m$ is then taken and is presented to the teacher as a batch for training. We train the teacher on the students most recent experience for $m$ steps and then train the teacher on the random sample of historic experience of size $m$. 

The architecture used for the teacher network is a fully-connected feedforward neural network with 3 hidden layers and 50 nodes in each layer. The hidden layer nodes are activated by the ReLU function and a dropout rate of 20\% is used to prevent overfitting. Stochastic gradient descent with momentum \cite{sutskever2013importance} is used to train the network weights with a momentum value of 0.9. A discount factor of 0.95 is used and the target network is updated after every 20 times the non-target network is updated. An experience replay batch size of 10 is used after every 10 updates on the non-target network. 

\subsection{The Teacher Action Space}
\label{section:teacher_action}
Maximum flexibility in data selection would be to allow for a continuous action space, such as making every entry of data in the training set a batch of data in itself. We use deep deterministic policy gradients as described in \cite{lillicrap2015continuous} which allows us to parameterise the action selection. With this approach, the teacher, $g$, consists of two networks, an actor and a critic. The actor is a network which selects the actions to take and the critic evaluates the `value' of being in a particular state. The actor, parameterised by $\theta^{\mu}$ (of which there is also a target actor), takes the encoded state of the student, $f$, as input and has two outputs. The first is the index along the curriculum (after ranking our training data according to the curriculum) at which we will centre our batch and the second is the width around this datapoint that we will expand our batch to. For example if our first output gives us the value 30 and the second output gives 50, then we will centre our curriculum batch on the datapoint in the curriculum indexed at 30 and select everything 25 below and 25 above this datapoint as a batch of data to train on. 
\par The critic, parameterised by $\theta^Q$ (of which there is also a target critic) has two outputs (for our two independent actions, index selection and batch width selection) which are the Q-values to which the reward is added for training. The input to the critic is the same as the input to the actor (the state of the student) except it is altered by concatenating onto this the actions selected by the actor which also have some exploration noise applied. The critic is then trained with loss as in Equation \ref{eq:Bellman_loss}. The actor is updated using the following loss:
\begin{equation}
\label{eq:DDPG_loss}
    \nabla \theta^{\mu} J \approx \frac{1}{N} \sum_i \nabla_a Q\left(s,a \given \theta^Q\right)\given_{s=s_i, a=\mu\left(s_i\right)} \nabla_{\theta^{\mu}} \mu\left(s\given \theta^{\mu}\right)
\end{equation}
where $Q$ is the Q-function and $\mu$ is the policy. For a derivation of this please see \cite{lillicrap2015continuous}. A pseudocode is included in Algorithm \ref{algorithim:ST_network} for a detailed description of how the teacher network is trained. 
\begin{algorithm}[h!]
 \KwData{Training dataset organised into $N$ batches of Mahalnobis curriculum }
\ initialise teacher critic network, $Q$, actor network, $\mu$\\
\ initialise target teacher and actor, $Q_{T}$, $\mu_{T}$\\
\ initialise random process $\mathcal{N}$ for action exploration\\
\ initialise replay buffer, $R$, select batchsize of replay, $m$\\
\ select update frequency value, $U$\\
\ select stable update value, $\tau$\\
\For{$x$ in $X$ students}{
\ initialise student network, $f_{x}$\\
\For{$i$ in $I$ iterations}{
{-Extract state of $f_{x}$, $s_i$}\\
-Select action $a_i = \mu \left(s_i\given\theta^\mu\right) + \mathcal{N}_i$ according to current policy and exploration noise\\
-Execute action $a_i$ and observe reward $r_i$ based on improvement in performance on validation set and observe new state $s_{i+1}$\\
-Store transition $\left(s_i, a_i, r_i, s_{i+1}\right)$ in replay buffer, $R$\\
-Sample random minibatch of $n$ transition tuples from $R$\\
-Set $y_i = r_i + \gamma Q_T\left(s_{i+1},\mu_T\left(s_{i+1}\given\theta^{\mu_T}\right)\given\theta^{Q_T}\right)$\\
-Update critic by minimising: \\
$L = \frac{1}{n}\sum_i\left(y_i - Q\left(s_i, a_i\given\theta^Q\right)\right)^2$\\

-Update actor using sampled policy gradient:\\
    $\nabla \theta^{\mu} J \simeq \frac{1}{N} \sum_i \nabla_a Q\left(s,a \given \theta^Q\right) \nabla_{\theta^{\mu}} \mu\left(s\given \theta^{\mu}\right)$\\

\eIf{$i$ mod U = 0}{
    Update target networks:\\ 
    $\theta^{Q_T} \leftarrow \tau\theta^Q + \left(1-\tau\right)\theta^{Q_T}$\\
    $\theta^{\mu_T} \leftarrow \tau\theta^\mu + \left(1-\tau\right)\theta^{\mu_T}$\\
    }{
    continue}

 }
 }
\caption{The student-teacher training routine for discrete batches using the DDPG algorithm}
\label{algorithim:ST_network}
\end{algorithm}
We also experiment with teachers with discrete action spaces and compare the performance. The discrete action space teachers are trained using more appropriate algorithms such as deep Q-networks (DQN). A pseudocode for this type of teacher can be found in the supplementary material.

\subsection{Student Network}
\label{section:student}
The state of the student network is defined to be the matrix of weights of each layer in the network i.e, $W^{ij}$, where $W^{ij}$ indicates the matrix of weights between layers $i$ and $j$. $W^{ij}$ is of size $M_i \times M_j$ where $M_i$ and $M_j$ are the number of nodes in layers $i$ and $j$ respectively. As the size of these matrices is dependent upon the number of nodes in each layer we seek a more compact representation of these matrices. We do this by comparing each row in a weight matrix to a fixed reference vector. For example, $W_{1:}^{ij} \cdot a$ would compare the first row of matrix $W^{ij}$ with our fixed reference, $a$. The purpose of the reference vector is to allow us to compare the vectors of weights and allow us to identify them relative to the same reference. We represent each row using two scalars, $\lvert \langle W_{n:}^{ij}, a \rangle \rvert$ and $\angle\left(W_{n:}^{ij}, a\right)$ for $n = 1,2, \dots , M_i$. These are concatenated to give a vector representing a weight matrix between two layers, denoted by $\mathbf{v}_i \in \mathbb{R}^{2M_i}$. This is repeated for the next layer, giving $\mathbf{v}_j \in \mathbb{R}^{2M_j}$ and the vectors for each layer are then also concatenated together as a single vector, $\mathbf{v} \in \mathbb{R}^{2\left(M_i + M_j\right)}$. This vector $\mathbf{v}$ is then used as the representation of the state of the student. As described in Section \ref{section:teacher}, the performance of this network on a separate validation set as well as the training set is used as the reward which is used to update the teacher. 

We define the student to be a fully-connected feedforward neural network with  2 hidden layers, with nodes $M_i = M_j = 50$.  Each node in the hidden layer is activated by the ReLU function apart from the final layer where a softmax function is used to classify. The student is trained using stochastic gradient descent with a fixed learning rate of 0.001. We initialise a student and allow it to be trained using $N=100$ separate batches sorted by the Mahalanobis distance. We then initialise another student and repeat the procedure. We repeat this for 10,000 students, therefore providing the teacher with 1,000,000 tuples in the experience replay buffer, $R$, (the collection of the agents experiences) to sample and train from. At test time we initialise another student and allow the teacher to implement the learned policy.

\section{Datasets}
\subsubsection*{Ward Admission Dataset}
In this study we considered the patient data collected in the electronic health records (EHR) of a hospital trust that the authors are associated with between January 2013 and April 2017. This includes administrative (e.g. date and time of arrival), demographic (e.g. age, gender and so on), as well as physiological and medical information (e.g. vital sign measurements and medical tests ordered during the patient's visit). We only consider patient arrivals to the emergency department. Any historical data stored about the patient will also be available in the EHR upon their next arrival to the emergency department. As there are over 100 potential locations of admission within the hospital, we group these into seven `ward functions' where wards in a single `function' are capable of treating patients with similar conditions making this a seven-class classification. Only patients who were admitted in an emergency were considered providing a dataset of 14,324 patients. A training set of 60\% of the dataset was used and was balanced, leaving 8,589 patients for training on. The validation set was 20\% of the dataset and testing was also 20\% and the classes were kept in the same distribution as the original dataset.
\subsubsection*{MIMIC-III Dataset}
To validate the efficacy of the methodology we implement the algorithm on another classification problem from the MIMIC-III dataset in the next section \cite{johnson2016mimic}. The patients for this dataset are also emergency patients only and have all features available. This provides us with a dataset of 8806 patients. These were split into the same train-validation-test proportions as before with only the training set being balanced as before.  As MIMIC-III is an ICU focused dataset, replicating the experiment we have carried out with the ward admission dataset is not possible. As a result we create a new problem of classifying the mortality of patients (binary classification) based on 11 features that are available early in the patient's admission. 
\subsubsection*{CIFAR-10}
To further validate our methodology we also report results on the CIFAR-10 image recognition dataset \cite{Krizhevsky09learningmultiple} to assess the model's performance on a non-tabular dataset. Due to our model having the capability of selecting large batches and memory constraints, we take a random, stratified sample from the training set of 12,000 images of which 1,200 are made into a validation set. We then take a random, stratified sample of 3000 images from the test set as our held-out test set. We repeat this three times essentially creating three separate datasets which we identify as sample 1, sample 2 and sample 3 respectively in Table \ref{table: MIMIC-III_results}.

\section{Results}
\label{section:results}
\begin{table*}[t!]\caption{Average classification accuracies and standard deviations for various baseline and state-of-the-art methods on the Ward Admission (tabular), MIMIC-III (tabular) and CIFAR-10 (image) datasets. All models are averaged over the same five seeds apart from those highlighted with * which indicates that the accuracy reported from the cited text is quoted.}
\vskip 0.15in
\begin{center}
\begin{small}
\begin{sc}
\begin{tabular}{lcccccr}
\toprule
       & Ward Admission & MIMIC-III & CIFAR-10 & CIFAR-10 & CIFAR-10\\
       &                &           & Sample 1 & Sample 2 & Sample 3\\
Method & Acc (SD) & Acc (SD) & Acc (SD) & Acc (SD) & Acc (SD)  \\
\midrule
Batchwise    & 0.45 (0.01) & 0.63 (0.01)& 0.65 (0.02)&0.65 (0.02)&0.65 (0.02)\\
Curriculum & 0.48 (0.02) & 0.63 (0.02)& 0.68 (0.01)&0.68 (0.01)&0.68 (0.01) \\ 
ST + curric    & 0.53 (0.03) & 0.63 (0.02)& 0.68 (0.02)&0.68 (0.02)&0.68 (0.02) \\
DeepFM    & 0.59 (0.01) & 0.66 (0.01)&  N/A & N/A & N/A       \\
Deep+CrossNet     & 0.58 (0.02) & 0.68 (0.02)& N/A & N/A & N/A\\
AutoInt      & 0.57 (0.02) & 0.67 (0.01)& N/A & N/A & N/A\\
DenseNet*    & N/A & N/A & 0.96 (0.01) & 0.96 (0.01) & 0.96 (0.01) \\
GPipe*      & N/A & N/A & \textbf{0.99} (0.01) & \textbf{0.99} (0.01) & \textbf{0.99} (0.01) \\
DQN student-teacher      & 0.58 (0.01) & 0.66 (0.02)& 0.86 (0.04) & 0.88 (0.02) & 0.86 (0.03) \\
DDPG student-teacher      & \textbf{0.62} (0.02) & \textbf{0.70} (0.01)& 0.90 (0.01) & 0.90 (0.01) & 0.89 (0.01) \\
\bottomrule
\end{tabular}
\end{sc}
\end{small}
\end{center}
\vskip -0.1in
\label{table: MIMIC-III_results}
\end{table*}

To make a comparison between our student-teacher network and other state-of-the-art methods for classification, we show normal stochastic mini-batch training (batchwise), vanilla curriculum training (curriculum) and the method proposed by \cite{matiisen2017teacher} (ST + curric). We report the (ST + curric) result for the student-teacher method that worked best for each dataset out of the four approaches discussed in Section \ref{section:relatedwork}. The performances for both methods of batch formation (as discussed in Section \ref{section:pre-processing}) were compared. It was found that for direct curriculum training (curriculum) method 2 was best whereas for the student-teacher methods (ST + curric, DQN student-teacher, DDPG student-teacher), method 1 worked best. 

For the Ward Admission and MIMIC-III datasets we also compare to state-of-the-art methods for classifying tabular data, namely DeepFM \cite{guo2017deepfm}, Deep+CrossNet \cite{wang2017deep} and AutoInt \cite{song2019autoint}. For the CIFAR-10 dataset we compare with models that achieve state-of-the-art performance on image datasets, namely DenseNet \cite{huang2017densely} and GPipe \cite{huang2019gpipe}. For our non-medical dataset, CIFAR-10, we sort data using cosine similarity as entropy. We must use a different student for this dataset and so we use a student that has 4 convolutional layers with 32 filters of size 3x3 in the first two layers and 64 of these in the second two. These are all activated by ReLU and maxpooled and are followed by 3 feedforward layers of size 50 nodes each. We use the encoding of the weights of the feedforward component of the network as the state of the student. From Table \ref{table: MIMIC-III_results}, we see that our method (DDPG trained student-teacher network) outperforms at least 3 state-of-the-art methods for classifying tabular data. It is also capable of providing a competitive performance on image classification problems, approaching state-of-the-art performance. This demonstrates the robustness of this method as a training methodology for various datasets. In Section \ref{section:discussion} and the supplementary material we go on to investigate the policies that have been generated for training and discuss them in the context of the curriculum.
\section{Investigation of Policies Learned}
\label{section:discussion}
\subsection*{Learned Curricula}
\begin{figure}[h!]
  \centering
  \begin{minipage}[b]{0.35\textwidth}
    \includegraphics[width=\columnwidth]{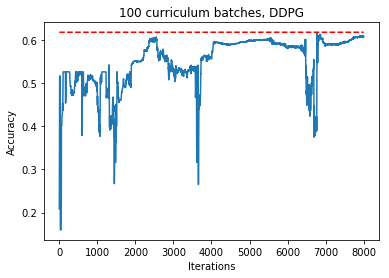}\caption{The performance of the student on the held-out test set of the Ward Admission prediction while it is trained by the teacher. The red dashed line is the best performance achieved by this student. Notice that it does not follow the typical smooth learning arc usually seen.}
    \label{fig:100batch_actor}
  \end{minipage}
  \hfill
  \begin{minipage}[b]{0.35\textwidth}
    \includegraphics[width=\columnwidth]{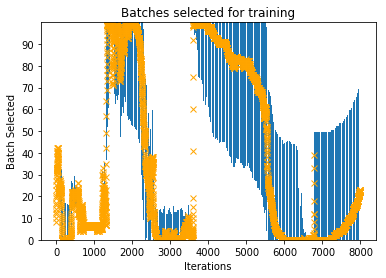}\caption{Actions generated by the learned policy of the teacher that has led to the performance of the student shown in Figure \ref{fig:100batch_actor}. Orange crosses are the first output (where to select data from) and blue bars are the second output (how much data around the central selection point to include in the batch for training). If the batch selected is near zero then this is low entropy data and if it is near the top of the batch selection then this is high entropy data.}
    \label{fig:100batch_policy}
  \end{minipage}
\end{figure}

In this section we investigate the policies learned by the teachers through their experience of training many students. We implement the student in such a way that the configuration of the student at which the highest accuracy is achieved is saved. From Figure \ref{fig:100batch_actor} we see that the student accuracy curve does not follow the usual smooth training arc seen during stochastic mini-batch training. We can see areas where the teacher aids the student in escaping local minima. The teacher swinging from training the student with low entropy data to high entropy data just before iteration 4000 in Figure \ref{fig:100batch_policy} corresponds to a degradation in performance of the student in Figure \ref{fig:100batch_actor}. This may indicate that the teacher has used the high entropy data to push the student out of a local minimum hence why the performance degrades. This turns out to be an unsuccessful attempt as with further training the model does not perform any better than it had done previously. However, just prior to iteration 7000, the teacher once again performs a `high entropy spike' (shown by the orange crosses rising). This again deteriorates the performance by escaping from a local minimum, but with further training the student achieves a performance of 62\% (indicated by the red dotted line in Figure \ref{fig:100batch_actor}, the highest it has achieved so far. We show policies learned for the MIMIC-III and CIFAR-10 datasets in the supplementary material.
\subsection*{Constrained Policy Learning}
To demonstrate that the curriculum is indeed guided by the teacher and not simply providing an alternative optimisation trajectory, we train two identical teachers, one with no constraints and one that has been constrained in some way and we compare the policies learned. We use a DDPG teacher where the constrained teacher uses 1 datapoint at a time with probability 0.999 (i.e., the teacher's second output (batch width) is almost always 0) and uses the real teacher output otherwise. The students of these teachers are trained on the CIFAR-10 dataset.
\begin{figure}[h!]

  \centering
  \begin{minipage}[b]{0.35\textwidth}
    \includegraphics[width=\columnwidth]{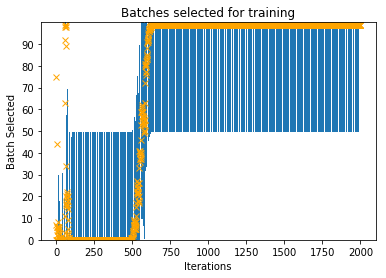}\caption{Actions generated by the learned policy of an unconstrained DDPG teacher on the CIFAR-10 dataset. Orange is the first output (where to select data from) and blue is the second output (how much data to include in the batch). If the batch selected is near zero then this is low entropy data and if it is near the top of the batch selection then this is high entropy data.}
    \label{fig:DDPG unconstrained policy}
  \end{minipage}
  \hfill
  \begin{minipage}[b]{0.35\textwidth}
    \includegraphics[width=\columnwidth]{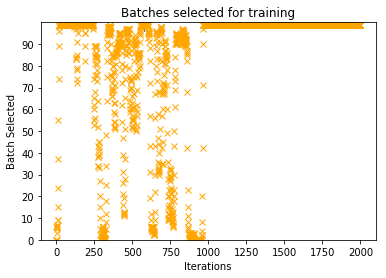}\caption{Actions generated by the learned policy of a constrained DDPG teacher on the CIFAR-10 dataset where it is highly probable that only one datapoint will be selected at a time for training. We see that the selection is significantly more chaotic. Batch numbers near zero are low entropy data and batch number 100 contains the highest entropy data.}
    \label{fig:DDPG constrained policy}
  \end{minipage}
\end{figure}
From Figure \ref{fig:DDPG unconstrained policy}, we see a fairly typical curriculum with a relatively noisy starting phase going from low to high entropy then rapidly back to low. However the general trend is that low entropy data is used and trained with before progressing onto the higher entropy data. In Figure \ref{fig:DDPG constrained policy} we see that we begin in a similar way as in Figure \ref{fig:DDPG unconstrained policy} (rapidly going from low to high entropy) but after this the smooth transition from low entropy to high entropy is not observed. It would seem that due to individual datapoints providing much noisier gradients than batch updates, the teacher must oscillate rapidly between high and low entropy data in an attempt to provide effective training. It is interesting to observe that the teacher in both cases samples along the entire entropy axis before converging to high entropy. We also show in the supplementary material the effect of reducing the learning rate of the student as another form of constraint.

\subsection*{Policy Stability for Similar Students}

\begin{figure}[h!]
  \centering
  \begin{minipage}[b]{0.35\textwidth}
    \includegraphics[width=\columnwidth]{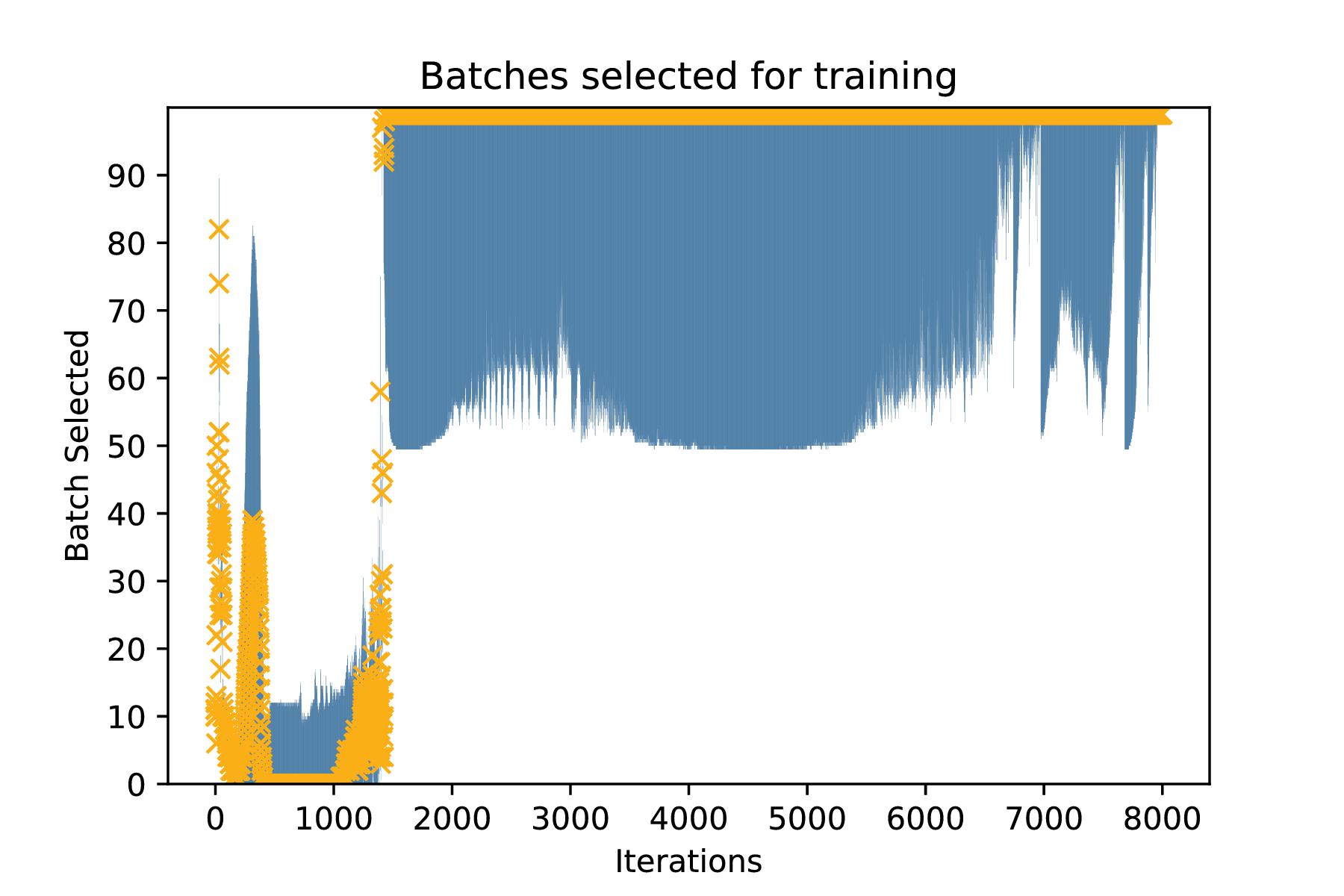}\caption{The actions generated by the teacher for a student to train on the Ward Admission dataset.}
    \label{fig:original_policy}
  \end{minipage}
  \hfill
  \begin{minipage}[b]{0.35\textwidth}
    \includegraphics[width=\columnwidth]{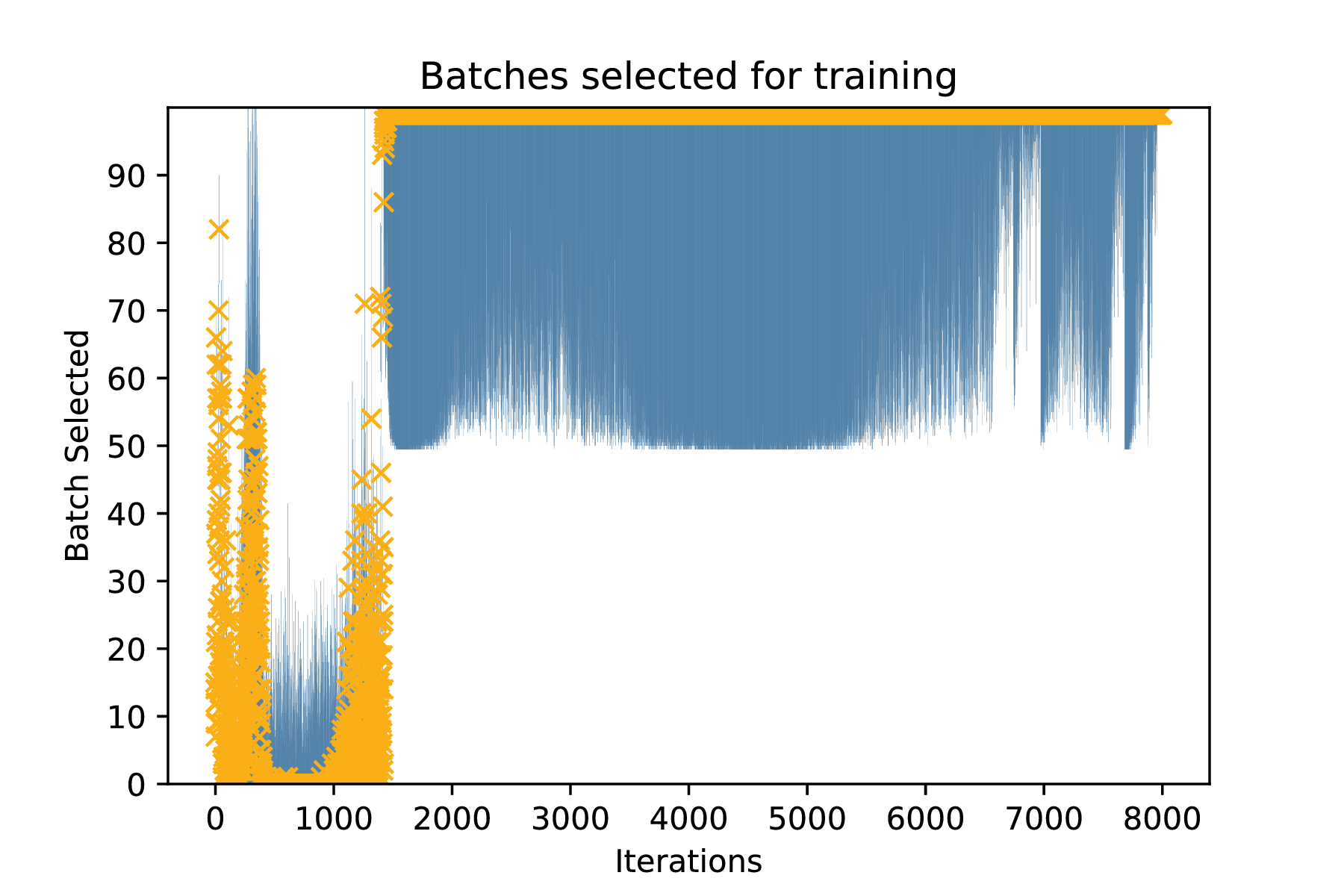}\caption{How the actions of the teacher shown in Figure \ref{fig:original_policy} have changed by applying Gaussian noise to the states of the student.}
    \label{fig:perturbed_policy}
  \end{minipage}
\end{figure}

To investigate the stability of the policies learned by the teacher we save the state of the student at every iteration and corrupt this signal with zero-mean Gaussian noise with standard deviation 0.1. We see from Figures \ref{fig:original_policy} and \ref{fig:perturbed_policy} that whilst the policies have changed slightly, the overall strategy of the policy is the same. This is encouraging as this is indicative of learned behaviour conditioned on the state as opposed to providing an alternative form of stochastic training. We provide further examples of stable policies for the other datasets in the supplementary material.

\subsection*{Policy Transfer between Tasks}

\begin{figure}[h!]
  
    \centering
  \begin{minipage}[b]{0.35\textwidth}
    \includegraphics[width=\columnwidth]{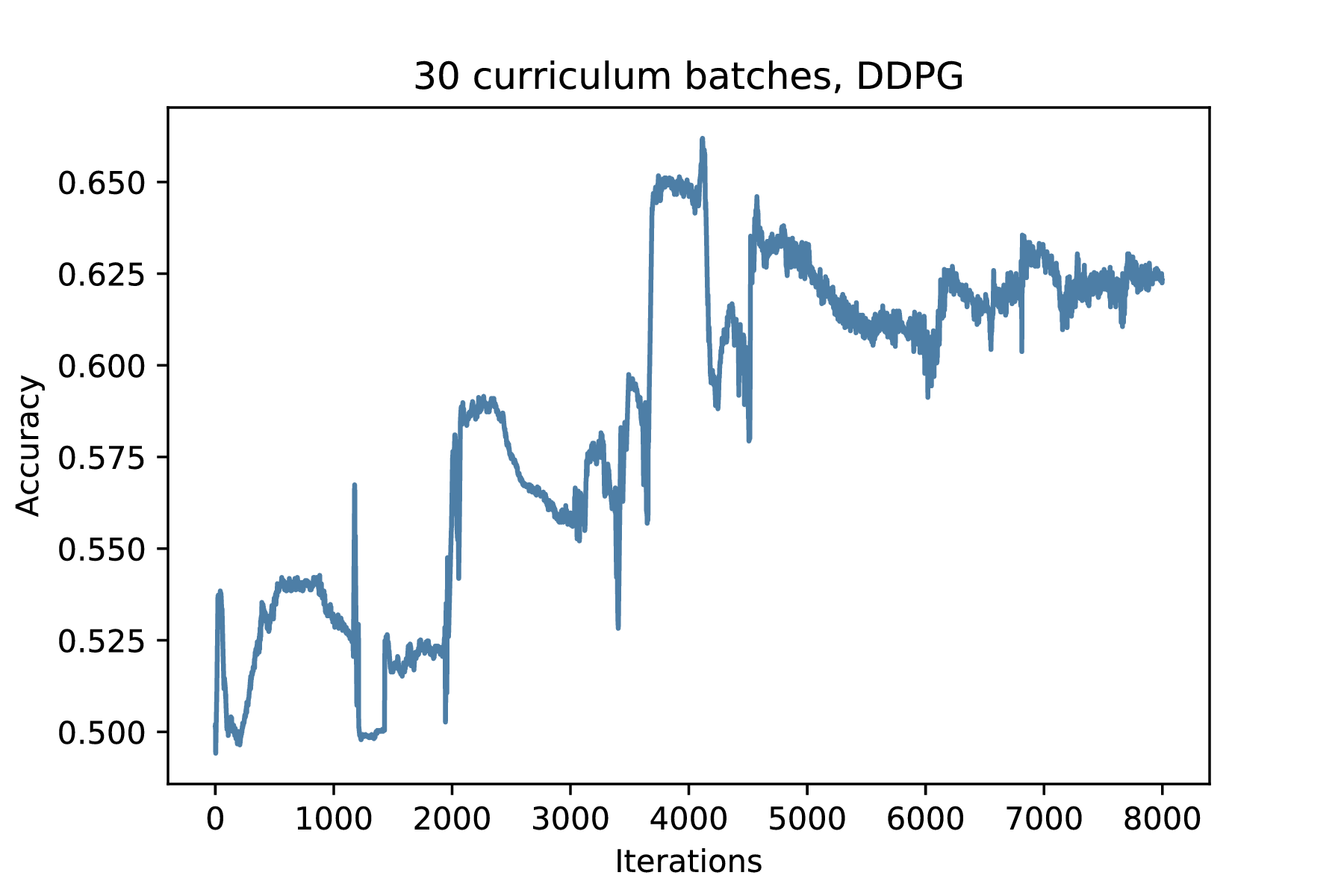}\caption{Performance of student on MIMIC-III dataset when trained by a transferred teacher taught to train students on the Ward Admission dataset.}
    \label{fig:transferred_actor}
  \end{minipage}
    \begin{minipage}[b]{0.35\textwidth}
    \includegraphics[width=\columnwidth]{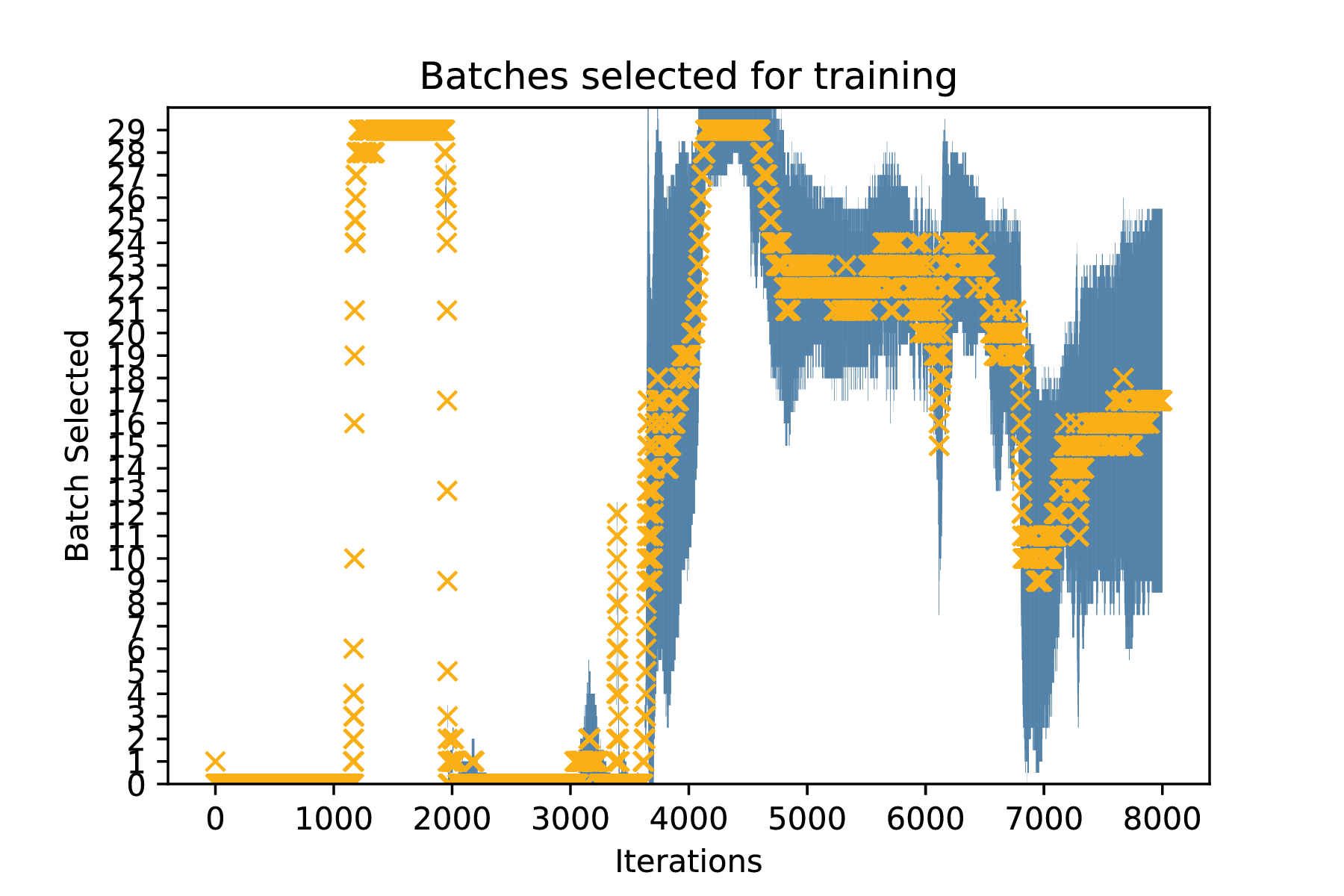}\caption{Actions generated by a transferred policy for the randomly initialised student which yielded the performance shown in Figure \ref{fig:transferred_actor}.}
    \label{fig:transferred_policy_mimic}
  \end{minipage}
\end{figure}

We also investigated how transferable the trained teachers are between tasks. A teacher was trained on the Ward Admission dataset and transferred to the MIMIC-III prediction problem with no fine-tuning. The policy the teacher generated on the `as before unseen' training set is shown in Figure \ref{fig:transferred_policy_mimic} and the corresponding result is shown in Figure \ref{fig:transferred_actor}.

Here we see that the teacher once again uses the `entropy spiking' technique to escape possible local minima resulting in the waves of improvement in accuracy shown in Figure \ref{fig:transferred_actor}, characteristic of curriculum learning. The performance attained is less than it would be when training the teacher directly using the MIMIC-III dataset (averaging 67\% $\pm$ 1\% versus 70\% $\pm$ 1\%, however it still outperforms the baseline methods and performs equivalently to some of the state-of-the-art methods (shown in Table \ref{table: MIMIC-III_results}). It is not currently clear why the transfer of teachers seems to be successful but this is being investigated for various datasets. More examples of policies generated by transferred teachers can be seen in the supplementary material.
\section{Conclusion and Limitations}
\label{section:future_work}
In this work we have presented a method of predicting which type of hospital ward a patient will be admitted to after being triaged in the ED. We believe that this can prove a useful tool for hospitals to improve the flow of patients out of the ED and into the hospital. We have also shown that a `teaching' neural network can learn from an encoding of the state of another network. In this way an appropriate curriculum can be generated by the former in order to maximise the performance of the latter on the task at hand. It is hoped that the insights gained from using this approach will help with curriculum design as well as provide insight into how learning occurs in neural networks.

One of the limitations we face with this approach is that the representation of the weights of the student is a form of dimensional reduction and therefore may not be the best representation of the weights of the network for learning from. To improve upon this we propose in future works to introduce an attention mechanism to identify which nodes in the network are predominantly responsible for the prediction. We will then use these weights as inputs to the teacher to remove redundancies from the input. There is also the problem that the number of inputs to the teacher relies on the size of the student where a student with $M$ nodes in $L$ hidden layers results in an input size of order $M$ for the teacher. While this is better than explicit representation of the weights which would result in an input size of order $M^2$, we actively seek for better representations of the student's state. One method of doing this is potentially through the use of distillation networks as seen in \cite{hinton2015distilling}, allowing for a more compact representation of the student network and then using the student-teacher methodology proposed in this work. We are encouraged however by the performance of the student-teacher setup on relatively small student networks achieving state-of-the-art performances. We also encourage experimentation with various curricula to rank data according to the dataset being considered. This changes the way data are ranked and therefore provides different combinations of data that can be grouped into batches. Alternatively we could investigate having different outputs for the teacher networks and investigate methods of combining these outputs to select data for training.






\nocite{langley00}

\newpage\phantom{blabla}
\newpage\phantom{blabla}

\appendix
\section{Examples of Policies Learned}

\subsection*{Learned Curricula}
Here we present some examples of the curricula that were learned by the teacher for the three datasets we have used. We show that the policies learned are consistent according to the dataset and reflect a strategy that has been learned by the teacher.

\subsubsection*{Ward Admission}

\begin{figure}[h!]
  \centering
  \begin{minipage}[b]{0.35\textwidth}
    \includegraphics[width=\columnwidth]{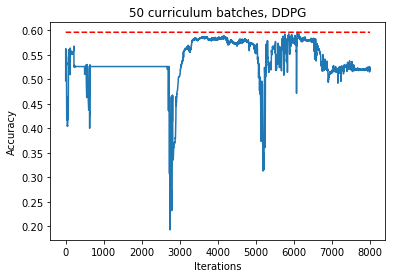}\caption{The performance of the student on the held-out test of the ward admission dataset while it is trained by the teacher.  The red dashed line is the best performance achieved by this student.}
    \label{fig:50batch_actor}
  \end{minipage}
  \hfill
  \begin{minipage}[b]{0.35\textwidth}
    \includegraphics[width=\columnwidth]{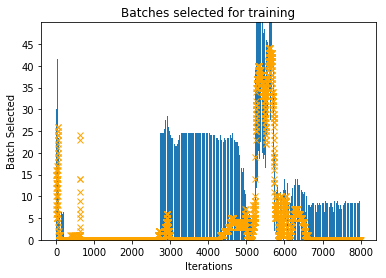}\caption{The actions generated by the policy of the teacher that has led to the performance of the student shown in Figure \ref{fig:50batch_actor}. Orange crosses are the first output (where to select data from) and blue bars are the second output (how much data around the central selection point to include in the batch for training). If the batch selected is near zero then this is low entropy data and if it is near the top of the batch selection then this is high entropy data.}
    \label{fig:50batch_policy}
  \end{minipage}
\end{figure}

We show another example of training by spiking in entropy to escape local minima in Figures \ref{fig:50batch_actor} and \ref{fig:50batch_policy}. Once again there is a spike in entropy of data selected for training prior to 6000 iterations, which allows us to escape a local minimum and degrade the performance but upon further training achieve a better accuracy on the held-out test set. It would seem that this entropy spiking strategy is the preferred strategy for the ward admission dataset.

\subsubsection*{MIMIC-III}
Plotted below are various examples of the curricula that were developed to train students on the MIMIC-III prediction problem. All of these provided state-of-the-art performance on the prediction problem.

\begin{figure}[h!]

  \centering
  \begin{minipage}[b]{0.35\textwidth}
    \includegraphics[width=\textwidth]{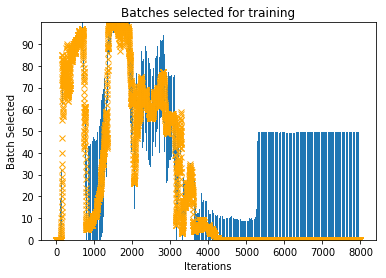}\caption{Curriculum generated for a randomly initialised student trained on the MIMIC-III dataset.}
    \label{fig:transferred_policy1}
  \end{minipage}
  \hfill
  \begin{minipage}[b]{0.35\textwidth}
    \includegraphics[width=\textwidth]{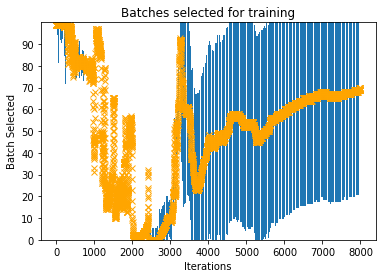}\caption{Curriculum generated for a randomly initialised student trained on the MIMIC-III dataset.}
    \label{fig:transferred_policy2}
  \end{minipage}
   \begin{minipage}[b]{0.35\textwidth}
    \includegraphics[width=\textwidth]{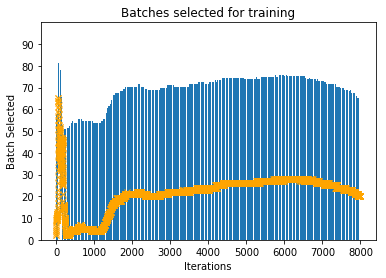}\caption{Curriculum generated for a randomly initialised student trained on the MIMIC-III dataset.}
    \label{fig:transferred_policy3}
  \end{minipage}
  
\end{figure}

In Figures \ref{fig:transferred_policy1} and \ref{fig:transferred_policy2} we see that the teacher utilises very small data batches to train. This generally gives rise to very noisy training gradients which it seems the teacher uses to converge to a favourable `initialisation' from which it then starts to train on bigger batches. In Figure \ref{fig:transferred_policy3} we see that the teacher seems to bring the student into a `good initialisation' early and so the rest of training is on the bigger batches.

\subsubsection*{CIFAR-10}

\begin{figure}[h!]
  \centering
  \begin{minipage}[b]{0.40\textwidth}
    \includegraphics[width=\columnwidth]{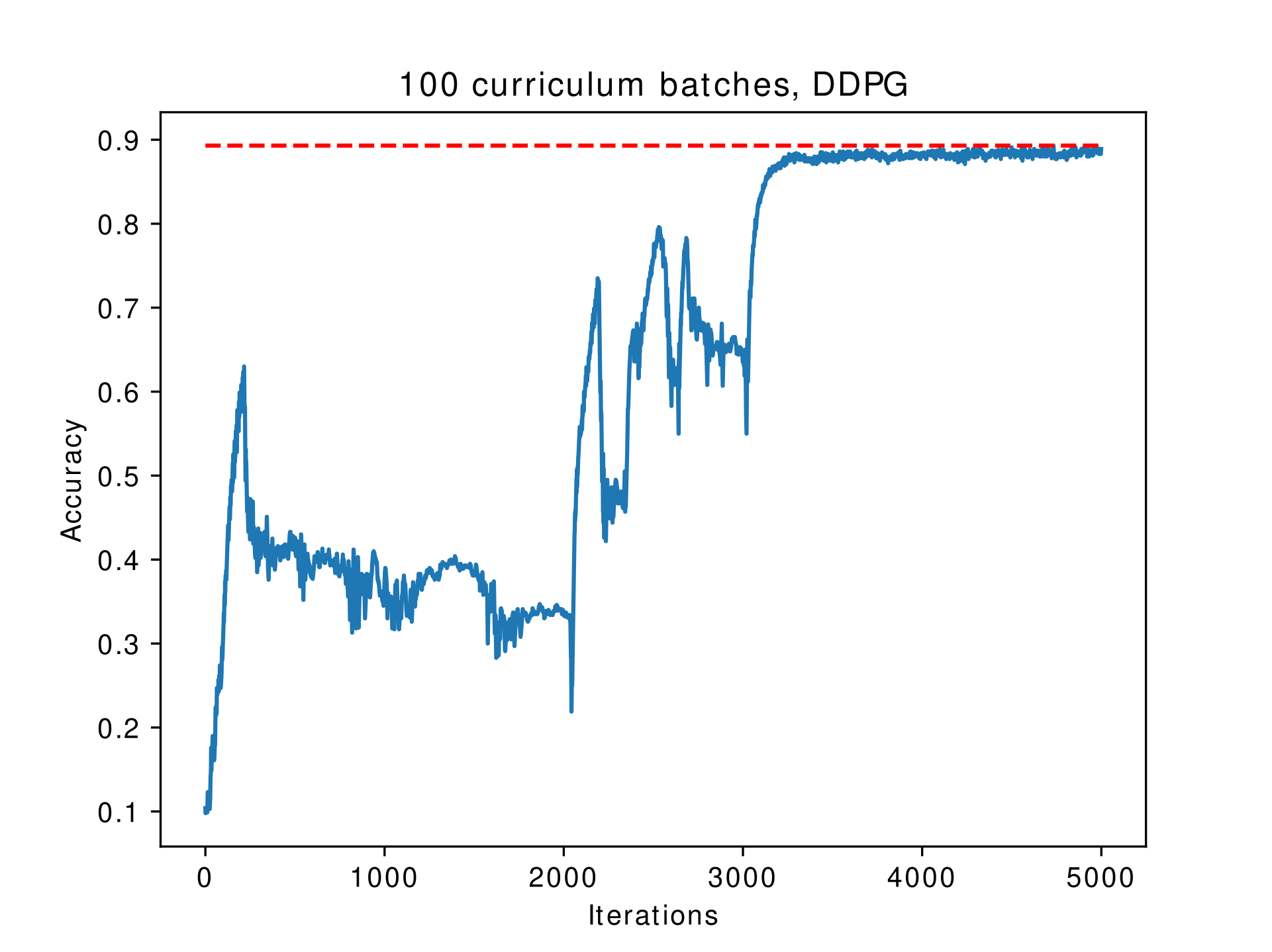}\caption{The performance of the student on the held-out test of the CIFAR-10 dataset while it is trained by the teacher.  The red dashed line is the best performance achieved by this student.}
    \label{fig:CIFAR_actor}
  \end{minipage}
  \hfill
  \begin{minipage}[b]{0.40\textwidth}
    \includegraphics[width=\columnwidth]{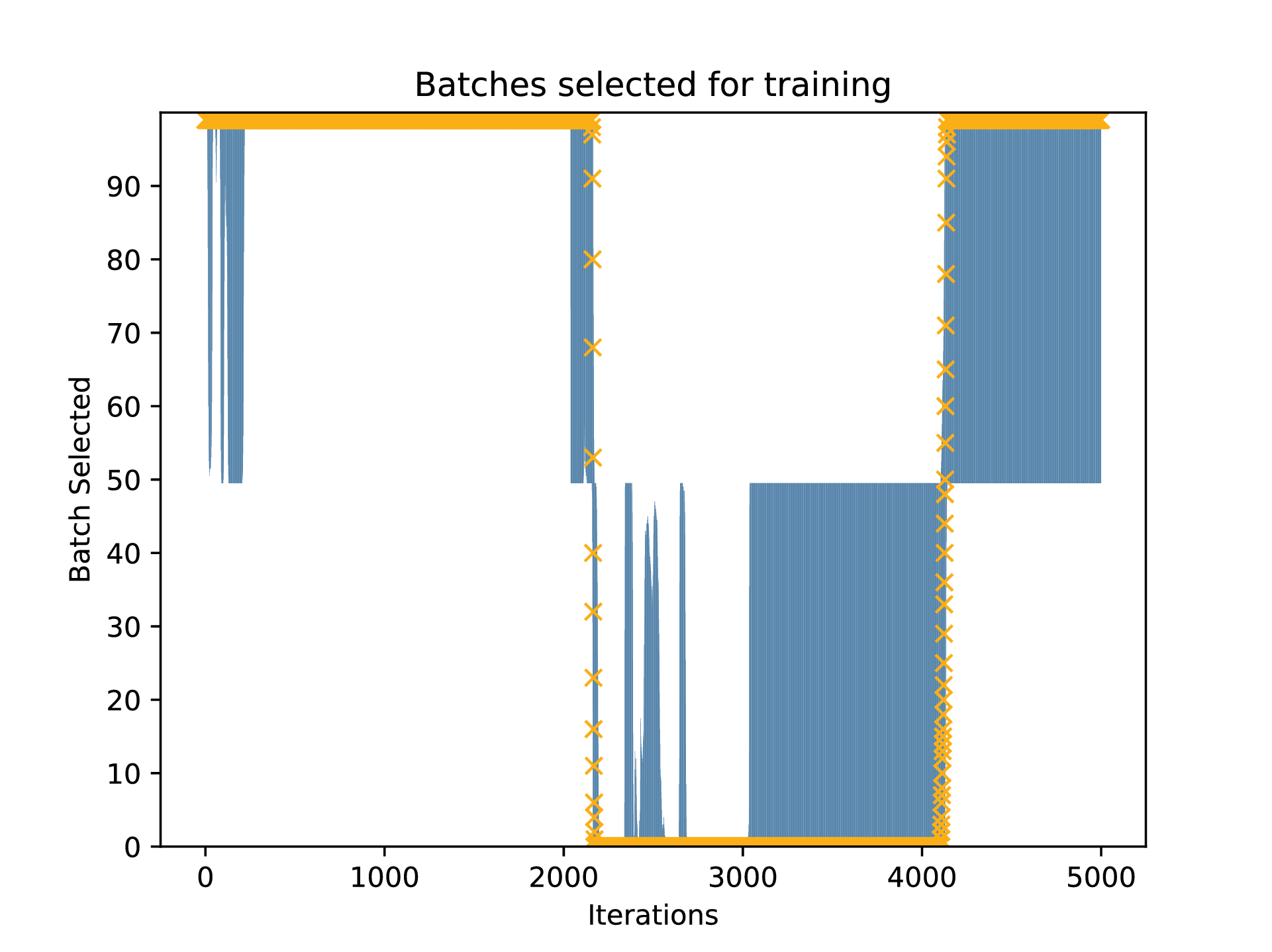}\caption{The actions generated by the policy of the teacher that has led to the performance of the student shown in Figure \ref{fig:CIFAR_actor}.}
    \label{fig:CIFAR_policy}
  \end{minipage}
\end{figure}

\begin{figure}[h!]
  \centering
  \begin{minipage}[b]{0.40\textwidth}
    \includegraphics[width=\columnwidth]{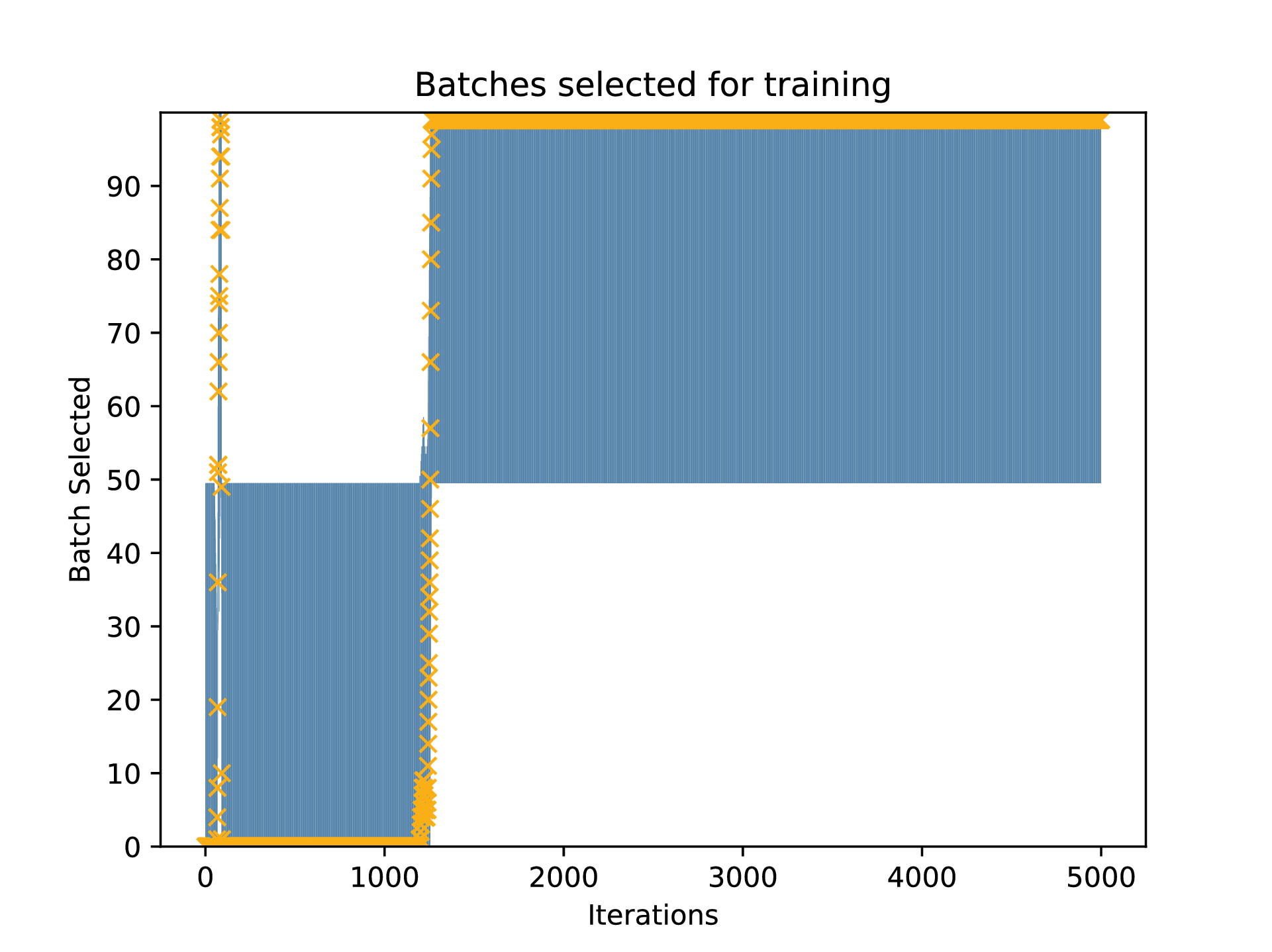}\caption{Curriculum generated for a randomly initialised student trained on the CIFAR-10 dataset.}
    \label{fig:CIFAR_policy1}
  \end{minipage}
  \hfill
  \begin{minipage}[b]{0.40\textwidth}
    \includegraphics[width=\columnwidth]{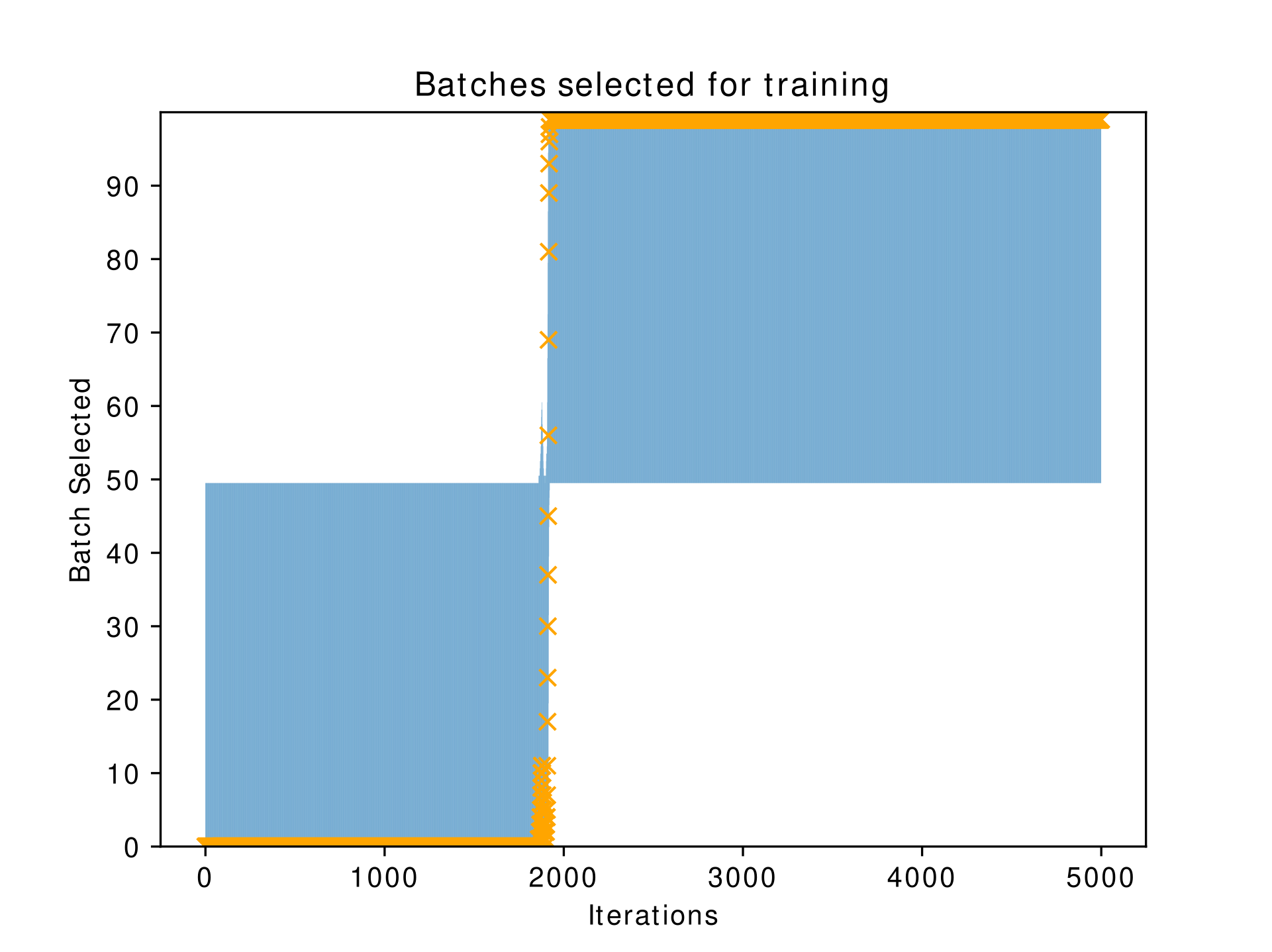}\caption{Curriculum generated for a randomly initialised student trained on the CIFAR-10 dataset. \ref{fig:50batch_actor}.}
    \label{fig:CIFAR_policy2}
  \end{minipage}
\end{figure}

The performance of a student and the curriculum learned for training this student on the CIFAR-10 dataset are shown in Figures \ref{fig:CIFAR_actor} and \ref{fig:CIFAR_policy}. We see the teacher primes the student into an initial state before (at approximately iteration 3000) repeatedly presenting low entropy batches before progressing to high entropy batches. This is very similar to curricula that are commonly used in many studies on image recognition. Figures \ref{fig:CIFAR_policy1} and \ref{fig:CIFAR_policy2} show the curricula used for other students by the same teacher. It would seem that repeated presentation of low entropy batches before progressing to repeatedly presenting high entropy batches is most beneficial for training the image recognition students. This makes sense due to the need for feature extraction in order to generalise to other images.

\subsection*{Constrained Policy Learning}
In this section we present our findings of the policies of the teacher networks on various students for different tasks. We present the findings on the CIFAR-10  dataset in the main paper and the findings on the MIMIC-III and Ward Admission datasets below.
\subsubsection*{MIMIC-III}

\begin{figure}[h!]

  \centering
  \begin{minipage}[b]{0.40\textwidth}
    \includegraphics[width=\textwidth]{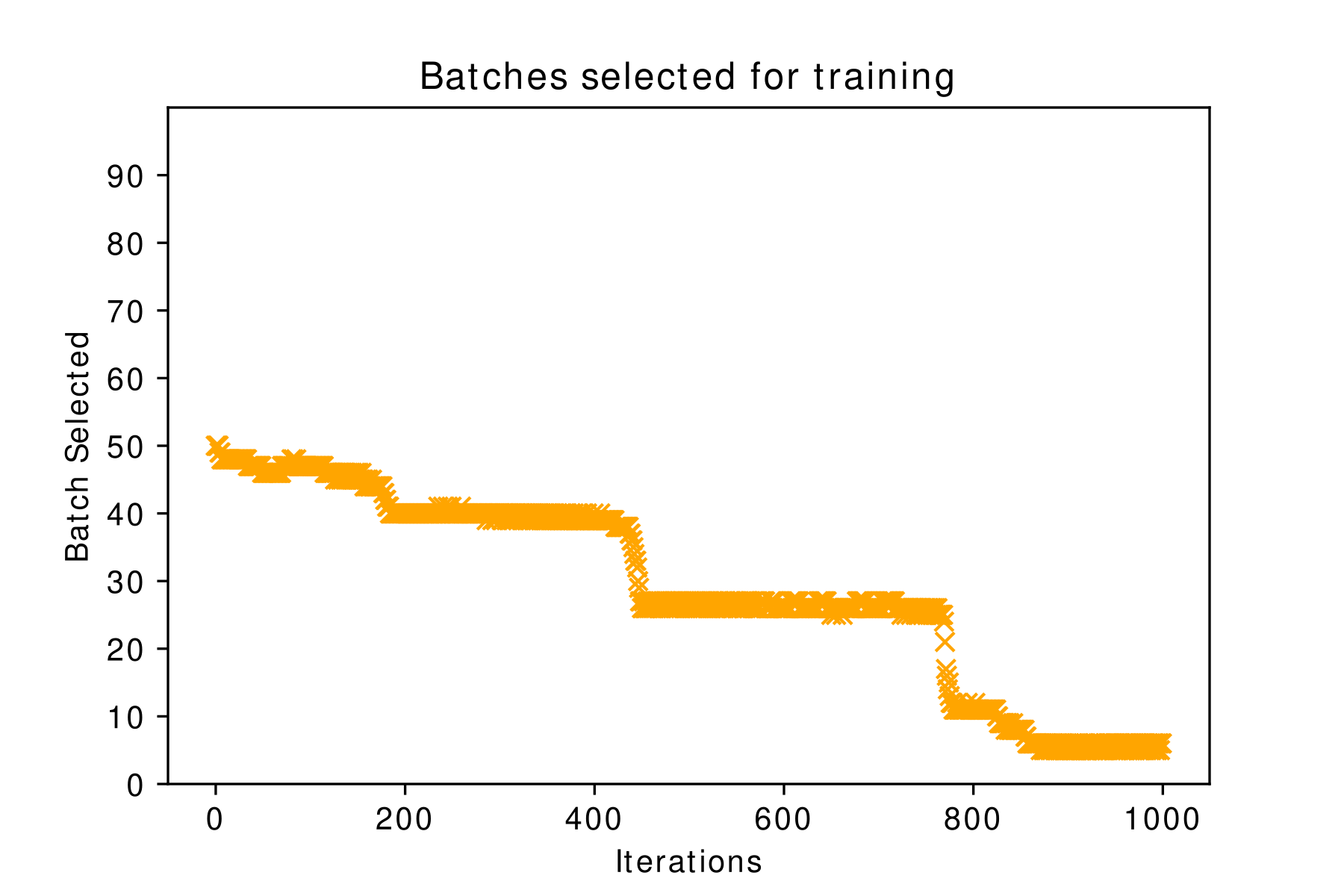}
    \caption{The actions generated by the learned policy of a constrained teacher to train a student on the MIMIC-III dataset. The student has a learning rate of 0.02.}
    \label{fig:mimic_fast}
  \end{minipage}
  \hfill
  \begin{minipage}[b]{0.40\textwidth}
    \includegraphics[width=\textwidth]{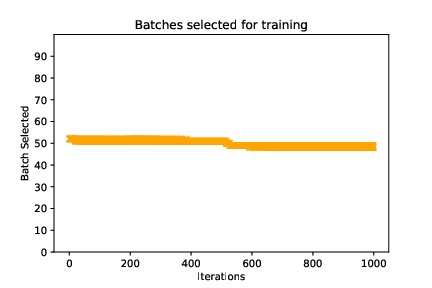}
    \caption{The actions generated by the policy of a constrained teacher to train a student that is also constrained with a lower learning rate of 0.002. The student has the same initial seed as that trained using the policy shown in Figure \ref{fig:mimic_fast}.}
    \label{fig:mimic_slow}
  \end{minipage}
\end{figure}

To constrain our students we first constrain our teacher (as done in the main paper) to select a batch width of zero with probability 0.999. Figure \ref{fig:mimic_fast} shows the policy of the teacher when training the student on MIMIC-III data. When comparing these to typical MIMIC-III generated curricula (Figures \ref{fig:transferred_policy1}, \ref{fig:transferred_policy2}, \ref{fig:transferred_policy3}), we see that there is no oscillation in entropy at the early stages of training and instead the teacher has learned to simply gradually step down in entropy. The student is trained with a learning rate of 0.02 and so in order to constrain this further we also reduce the students learning rate to 0.002, now constraining the student. We see from Figure \ref{fig:mimic_slow} that the teacher begins training the student using similar data (at approximately batch 50 on the entropy scale), however due to the student's lower learning rate the downward stepping takes significantly longer. This is highly encouraging as it shows that the teacher is following the same strategy as used in Figure \ref{fig:mimic_fast} albeit over a longer number of iterations as we would expect.

\subsubsection*{Ward Admission}

\begin{figure}[h!]

  \centering
  \begin{minipage}[b]{0.40\textwidth}
    \includegraphics[width=\textwidth]{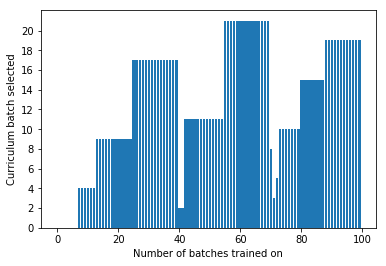}
    \caption{Actions generated by a DQN teacher with a learning rate of 0.01. At each iteration, anything shaded in blue is included in the batch used for training. }
    \label{fig:DQN unconstrained policy}
  \end{minipage}
  \hfill
  \begin{minipage}[b]{0.40\textwidth}
    \includegraphics[width=\textwidth]{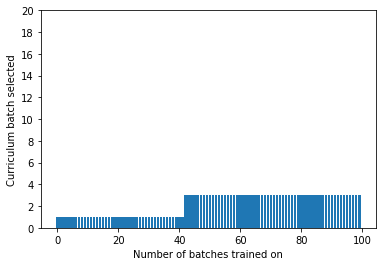}
    \caption{Actions generated by a DQN teacher with a learning rate of 0.001.}
    \label{fig:DQN constrained policy}
  \end{minipage}
\end{figure}

In Figures \ref{fig:DQN unconstrained policy} and \ref{fig:DQN constrained policy} we utilise a DQN trained teacher on the Ward Admission dataset. We initially train normally and then slow the learning rate of the student by 100 times for the same initial seed to see how this alters training.

We can see in Figure \ref{fig:DQN unconstrained policy} that a `recurring low to high entropy' curriculum is implemented by the teacher as seen implemented by the DDPG teacher. These can be seen as the DQN equivalent of the high entropy spiking strategy found by the DDPG teacher. Where we see drops in the entropy of data being used seem to be locations where the teacher is attempting to escape local minima. In Figure \ref{fig:DQN constrained policy} we reduce the student's learning rate and we see that we still have a `low to high entropy' curriculum but it is progressing much more slowly. Once again, this is due to the step size being smaller and therefore requiring more gradient updates to get the student network into a weight state that requires different batches for training.

\subsection*{Policy Stability for Similar Students}

We demonstrate in this section that for all the tasks considered our teacher learned stable policies conditioned on the current state of the student. We present our findings on the Ward Admission dataset in the main paper and our findings on the MIMIC-III and CIFAR-10 datasets below.

\subsubsection*{MIMIC-III}

\begin{figure}[h!]

  \centering
  \begin{minipage}[b]{0.40\textwidth}
    \includegraphics[width=\textwidth]{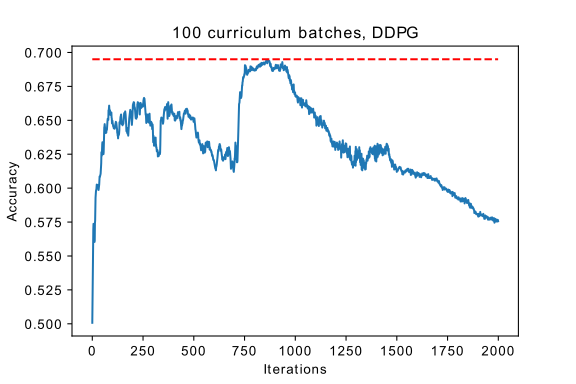}
    \caption{The performance of a student trained by the DDPG teacher on the MIMIC-III dataset. }
    \label{fig:mimic_nonperturbed_accuracy}
  \end{minipage}
  \hfill
  \begin{minipage}[b]{0.40\textwidth}
    \includegraphics[width=\textwidth]{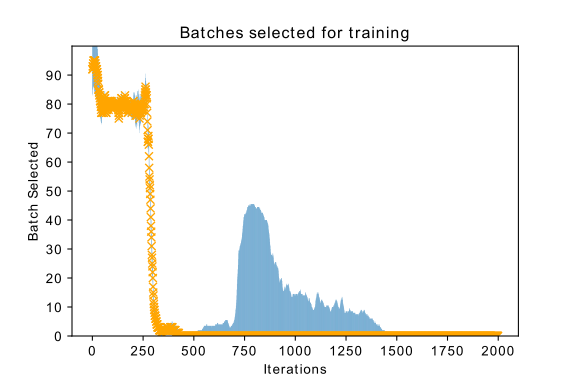}
    \caption{The actions used by the teacher to train the student with performance shown in Figure \ref{fig:mimic_nonperturbed_accuracy}.}
    \label{fig:mimic_nonperturbed_policy}
  \end{minipage}
\end{figure}

We see that once again the teacher learns a policy of using low entropy data to initialise the student before increasing the size of the batch introduced to maximise performance. We now once again apply Gaussian noise to the states of the student as done in the main paper.

\begin{figure}[h!]

  \centering
  \begin{minipage}[b]{0.40\textwidth}
    \includegraphics[width=\textwidth]{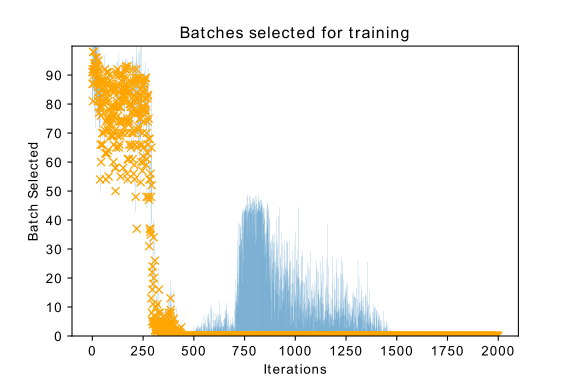}
    \caption{The actions taken by the teacher when the student has Gaussian noise applied to its states.}
    \label{fig:mimic_perturbed_policy}
  \end{minipage}
\end{figure}

In Figure \ref{fig:mimic_nonperturbed_policy} we see that the overall structure of the curriculum is the same as other MIMIC-III policies generated, beginning at high entropy and reducing to low to initialise the student before expanding the size of the batch. Figure \ref{fig:mimic_perturbed_policy} also shows this with a very similar curriculum to the one in Figure \ref{fig:mimic_nonperturbed_policy} being followed. This further encourages us that a strategy has indeed been learned by the teacher to train a student on the MIMIC-III dataset based on the weights of the student.

\subsubsection*{CIFAR-10}

Once again we repeat the exercise on the CIFAR-10 dataset and observe the stability of the teaching policy based on the corrupted states of the student.

\begin{figure}[h!]

  \centering
  \begin{minipage}[b]{0.40\textwidth}
    \includegraphics[width=\textwidth]{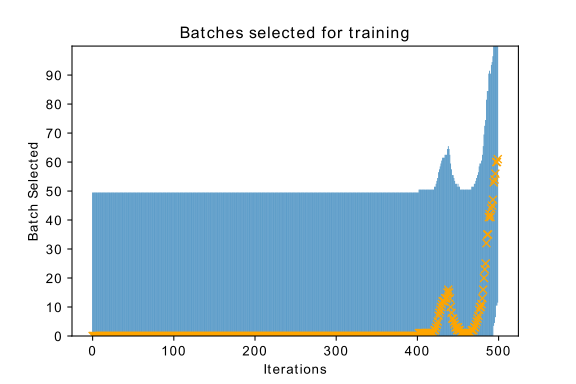}
    \caption{The actions used by the DDPG teacher to train a student on the CIFAR-10. }
    \label{fig:cifar_uncorrupted}
  \end{minipage}
  \hfill
  \begin{minipage}[b]{0.40\textwidth}
    \includegraphics[width=\textwidth]{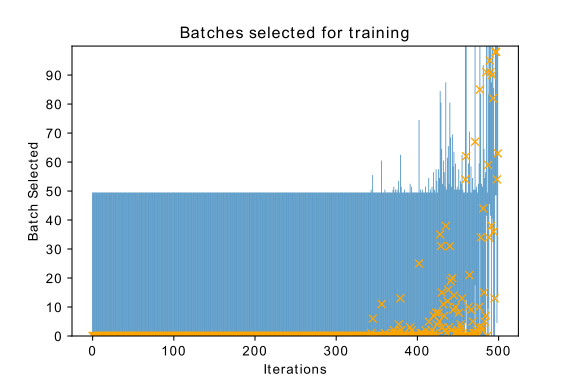}
    \caption{The actions of the teacher when the student has Gaussian noise applied to its states.}
    \label{fig:cifar_corrupted}
  \end{minipage}
\end{figure}

We see in Figures \ref{fig:cifar_uncorrupted} and \ref{fig:cifar_corrupted} that the same general policy is followed as that used in Figures \ref{fig:CIFAR_policy}, \ref{fig:CIFAR_policy1} and \ref{fig:CIFAR_policy2}. As we only train for 500 iterations the policy ends at the point of transition to training on high entropy data. We see that corrupting the students states with Gaussian noise has not significantly changed the policy of the teacher, providing further reassurance that the policy is not only stable but a learned function of the state of the student and not simply an alternative optimisation trajectory. 

\subsection*{Policy Transfer between Tasks}
In this section we provide further examples of policies generated from a teacher trained using the Ward Admission dataset on the MIMIC-III mortality prediction task.

\begin{figure}[h!]

  \centering
  \begin{minipage}[b]{0.40\textwidth}
    \includegraphics[width=\textwidth]{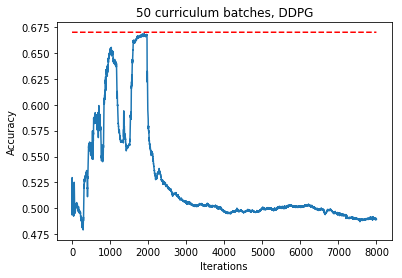}
    \caption{Performance of a randomly initialised student on the MIMIC-III dataset when trained by a teacher transferred from the Ward Admission dataset. }
    \label{fig:mimic_trans_acc}
  \end{minipage}
  \hfill
  \begin{minipage}[b]{0.40\textwidth}
    \includegraphics[width=\textwidth]{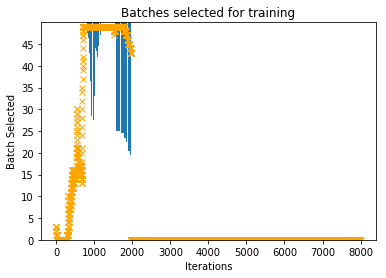}
    \caption{The actions selected by the transferred teacher when training the student for the MIMIC-III task.}
    \label{fig:mimic_trans_curric}
  \end{minipage}
\end{figure}

\begin{figure}[h!]

  \centering
  \begin{minipage}[b]{0.40\textwidth}
    \includegraphics[width=\textwidth]{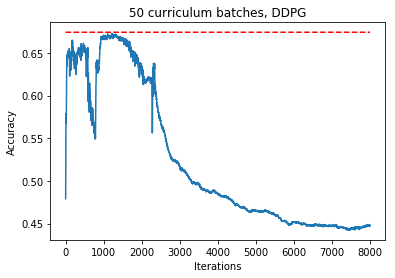}
    \caption{Performance of a randomly initialised student on the MIMIC-III dataset when trained by a teacher transferred from the Ward Admission dataset. }
    \label{fig:MIMIC_trans_Acc}
  \end{minipage}
  \hfill
  \begin{minipage}[b]{0.40\textwidth}
    \includegraphics[width=\textwidth]{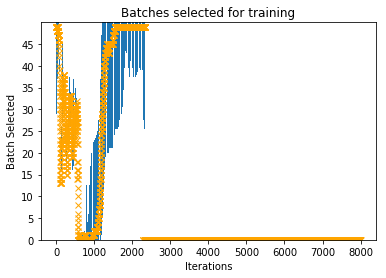}
    \caption{The actions selected by the transferred teacher when training the student for the MIMIC-III task.}
    \label{fig:MIMIC_trans_curric}
  \end{minipage}
\end{figure}

Figures \ref{fig:mimic_trans_acc} and \ref{fig:MIMIC_trans_Acc} show the performances of two randomly initialised students and the corresponding curricula that generated these performances are found in Figures \ref{fig:mimic_trans_curric} and \ref{fig:MIMIC_trans_curric} respectively. We see from these policies that the teacher uses the same strategy of small batches for initialisation and then the `batch expansion' as was seen when discussing the policies of the teacher trained using MIMIC-III. It is interesting to see that the teacher for Ward Admission also demonstrates this behaviour, however it is not clear why this is the case. Future work will investigate how we can characterise prediction problems in such a way that it is clear that the same teacher will generate appropriate curricula for them. We will also investigate how to combine teachers to train tasks that may be combinations of tasks or hybrid tasks and assess the curricula generated from these. We may also make the problem hierarchical, with a principal assigning teachers or combinations of teachers to train various students on tasks which can be ranked according to some metric (such as a task embedding). This metric can then be related back to the specialties of the teachers, with the principal using this information to use multiple teachers (one iteration at a time) or combinations of teachers to train the student on the task.




\clearpage
\section{Pseudocode for DQN Teacher}
\label{DQN_pseudo}
\begin{algorithm}[h!]
 \KwData{Training dataset organised into $N$ batches of Mahalanobis curriculum }
\ initialise teacher network, $g$\\
\ initialise target teacher by copying predictor teacher, $g^{T}$\\
\ select value of frequency of target network update and batchsize of replay data, $M$\\
\For{$x$ in $X$ students}{
\ initialise student network, $f_{x}$\\
\For{$i$ in $I$ iterations}{
{Extract state of $f_{x}$, $s$}\\
  \eIf{$i$ = 0}{
   train student on random batch (action), $a$\;
   }{
   select $a$ with highest Q-value from $g(s)$ according to a linearly decaying $\epsilon$-greedy policy with respect to $I$\;
  }
 \textbullet{train student ($f_{x}$) on action selected}\\
\textbullet{record performance improvement of student on training set and validation set and multiply for overall reward, $r$}\\
\textbullet{add $r$ to the output of $g^{T}(s)$ corresponding to the action taken to achieve this reward}\\
\textbullet{use the error between outputs of $g$ and $g{T}$ to backpropagate over the weights of $g$}\\
\textbullet{save $s$, $a$, $r$ and next state, $s'$ into replay buffer}\\
\eIf{$i$ mod $M$ = 0}{
    sample $M$ samples from replay buffer to train $g$ on\\
    update $g^{T}$ with new state of $g$\
    }{
    continue}

 }
 }
 \caption{The student-teacher training routine for discrete batches using the DQN algorithm}
 \label{algorithim:DQN_ST_network}
\end{algorithm}



%


\end{document}